%% file: sentaugment.tex
\documentclass[11pt,a4paper]{article}
\usepackage[hyperref]{acl2020}
\usepackage{times}
\usepackage{latexsym}

\usepackage{microtype}
\usepackage{graphicx}
\usepackage{subfigure}
\usepackage{booktabs} % for professional tables
\usepackage{url}
\usepackage{times}
\usepackage{amsmath}
\usepackage{latexsym}
\usepackage{array}
\usepackage{adjustbox}
\usepackage{multirow}
\usepackage{longtable}

\input{content/tables}

\aclfinalcopy % Uncomment this line for the final submission
\def\aclpaperid{479} %  Enter the acl Paper ID here

\usepackage{xspace}

\title{Self-training Improves Pre-training for Natural Language Understanding}

\author{Jingfei Du$^{\dagger}$\thanks{\ \ Equal contribution.} \space\space\space
  Edouard Grave$^{\dagger}$ \space\space\space
  Beliz Gunel$^{\ddagger}$ \space\space\space 
  Vishrav Chaudhary$^{\dagger}$ \space\space\space \AND
  \bf Onur Celebi$^{\dagger}$ \space\space\space 
  Michael Auli$^{\dagger}$ \space\space\space
  Ves Stoyanov$^{\dagger}$ \space\space\space
  Alexis Conneau$^{\dagger}$\footnotemark[1] \space\space\space \\ \\ \\
  $^{\dagger}$Facebook AI, $^{\ddagger}$Stanford University
  }
  
\date{}

\begin{document}
\maketitle
\begin{abstract}
Unsupervised pre-training has led to much recent progress in natural language understanding. In this paper, we study self-training as another way to leverage unlabeled data through semi-supervised learning. To obtain additional data for a specific task, we introduce SentAugment, a data augmentation method which computes task-specific query embeddings from labeled data to retrieve sentences from a bank of billions of unlabeled sentences crawled from the web. Unlike previous semi-supervised methods, our approach does not require in-domain unlabeled data and is therefore more generally applicable. Experiments show that self-training is complementary to strong RoBERTa baselines on a variety of tasks. Our augmentation approach leads to scalable and effective self-training with improvements of up to 2.6\% on standard text classification benchmarks. Finally, we also show strong gains on knowledge-distillation and few-shot learning.
%\footnote{Code will be available at \url{https://github.com/facebookresearch/SentAugment}}
\end{abstract}

\section{Introduction}

Self-training is a semi-supervised method which uses a teacher model, trained using labeled data, to create synthetic labels for unlabeled examples~\citep{scudder1965probability,yarowsky1995unsupervised}.
These synthetic labels are then used to train a student model.
This approach is called self-training when the student model has a similar or higher capacity than the teacher, and knowledge distillation~\citep{hinton2015distilling} when the student model is smaller than the teacher.
Self-training has been successfully applied to a variety of tasks, including image recognition~\citep{yalniz2019billion,xie2020self,zoph2020rethinking}, automatic speech recognition~\citep{synnaeve2019end,kahn2020self,park2020improved}, sequence generation~\citep{he2019revisiting}, and parsing~\citep{mcclosky2006parsing}.

An alternative semi-supervised technique is pre-training~\citep{dai2015semi,radford2018improving,howard2018universal,devlin2018bert}, which has led to large improvements for natural language understanding compared to purely supervised learning.
In that case, models are first trained on an auxiliary task, such as language modeling, followed by fine-tuning on the task of interest.

A natural question is the following: \emph{do pre-training and self-training capture the same information, or are they complementary?}
Recently, \citet{zoph2020rethinking} studied this question in the context of image recognition, showing that self-training was helpful, even in addition to pre-training.
However, their study mostly considers supervised pre-training, in which models were trained on ImageNet classification.
Moreover, in cases where large amounts of supervised data were available for the downstream task, pre-training was not helpful, even without self-training.
This is in contrast to natural language understanding for which language modeling pre-training is a very strong baseline that leads to large improvements for all the tasks we consider.

An important ingredient for self-training, and semi-supervised learning in general, is the unannotated data and the fact that it comes from the same domain as the downstream task.
Existing work, such as UDA~\citep{xie2019unsupervised}, self-training~\citep{he2019revisiting,xie2020self} and back-translation for machine translation~\citep{bojar2011bt_pbmt,sennrich2015improving,edunov2018bt}, assumes the existence of unannotated data in the same domain as the downstream task. This assumption limits the broad application of such semi-supervised methods, in particular in the case of low-resource downstream tasks. A second important question is thus: \emph{how can we obtain large amounts of unannotated data from specific domains?}

In this paper, we propose a data augmentation method, SentAugment, to build datasets of ``in-domain'' data for a given task from data crawled on the web. Web data covers many domains, and is available in large quantities. We use a large bank of web documents and construct sentence embeddings~\citep{kiros2015skip,wieting2015towards,conneau-EtAl:2017:EMNLP2017,artetxe2018massively,cer2018universal,arora2019simple}
that allow us to retrieve domain-specific unannotated sentences, which are similar to the existing training set of the downstream tasks. Our sentence embedding model 
%\beliz{It might be better to phrase it as "We adopt the paraphrastic sentence embedding model proposed by Wieting et al. instead of saying "Our paraphrastic sentence embedding model"}
is optimized for similarity search, trained with a triplet loss on ground-truth paraphrases, parallel sentences as well as as hard negatives~\citep{wieting2015towards,wieting2017paranmt}. 
We train a teacher model using the labeled task data and then further use it to synthetically label the retrieved sentences, and train the final model based on this synthetic dataset. Experiments show that SentAugment is effective for self-training, knowledge distillation and few-shot learning. The approach is generally applicable to new problems, leading to improvements on a variety of domains and tasks such as hate-speech and movie review classification over a strong RoBERTa~\citep{devlin2018bert,roberta2019} baseline.
To the best of our knowledge, this is the first study showing that self-training is complementary to a strong pre-training baseline for natural language understanding.
%\beliz{It would be helpful to have a contributions paragraph, currently it's slightly confusing for the reader to follow what our full approach is and what the contributions are on top of existing work.}
Specifically, we make the following contributions:
\begin{itemize}
    \item We introduce SentAugment, a data augmentation approach for semi-supervised learning that retrieves task-specific in-domain data from a large bank of web sentences. 
%crawled from the web.
    \item We show that self-training improves upon unsupervised pretraining: we improve RoBERTa-Large by 1.2\% accuracy on average on six standard classification benchmarks.
    \item We show that self-training improves accuracy by 3.5\% on average for few-shot learning.
    \item For knowledge-distillation, our approach improves the distilled RoBERTa-Large by 2.9\% accuracy on average, reducing the gap between the teacher and the student model.
    \item We will release code and models for researchers to build on top of our work.

\end{itemize}

\insertapproach

\section{Approach}
\label{sec:model+data}
Our SentAugment approach retrieves task-specific in-domain unsupervised data from a large bank of sentences which is used for self-training, where the teacher model - a RoBERTa-Large model finetuned on the downstream task - synthetically labels it. The synthetic labeled data is finally used to train the output student model (see Figure~\ref{fig:sentaugment}). We give more details on our approach in what follows.

\subsection{SentAugment: data augmentation for semi-supervised learning}
Whereas most semi-supervised approaches rely on in-domain unlabeled data, we are constructing similar datasets on the fly from the large bank of unannotated text. In what follows, we describe our data retrieval strategy for augmentation.

\paragraph{Large-scale sentence bank.}
Our approach relies on a large-scale corpus of unsupervised sentences, derived from data crawled on the web~\citep{wenzek2019ccnet}.
Because of its scale and diversity, our sentence bank contains data from various domains and with different styles, allowing to retrieve relevant data for many downstream tasks.
We embed each sentence using a universal paraphrastic sentence encoder~\citep{wieting2015towards,arora2019simple,ethayarajh2018unsupervised}, a model which was trained to output similar representations for sentences of similar meaning.
This sentence embedding space does not depend on the downstream tasks, and will be used to retrieve subsets of the sentence bank which are relevant to particular tasks.
For sentence encoders, we consider word2vec embeddings~\citep{mikolov2013word2vec,mikolov2018advances} and uSIF~\citep{ethayarajh2018repnlp}. We also train our own English sentence encoder, a Transformer pretrained with masked language modeling and finetuned to maximize cosine similarity between similar sentences. Specifically, we use a triplet loss $\mathcal{L}(x,y) = max(0, \alpha - cos(x,y) + cos(x,y_c))$ where positive pairs $(x,y)$ are either paraphrases or parallel sentences~\citep{wieting2019simple} and $y_c$ are in-batch hard negatives~\citep{wieting2015towards}.

\paragraph{Downstream task embeddings.}
For each downstream task, we build embeddings that are representative of the task, using the same paraphrastic model.
Then, we use these \emph{task embeddings} as queries for retrieving similar sentences from the sentence bank, using cosine similarity in the embedding space.
Specifically, we consider three ways for computing the task embeddings:
\emph{all-average}, where we obtain one embedding by averaging the sentence embeddings of all the samples from the training set of the downstream task ;
\emph{label-average}, where we construct one embedding per label, corresponding to the average of the sentence embeddings in the train set for each label ;
\emph{per-sentence}, where we keep one embedding for each sentence on the training set of the downstream task.

\paragraph{Unsupervised data retrieval.}
Using task-representative embeddings as queries, we retrieve a subset of our large sentence bank, corresponding to a few million sentences which we use as in-domain candidates for semi-supervised learning.
Reducing the amount of unannotated data is an important step as synthetically annotating billions of sentences using a large Transformer does not scale. We perform additional filtering based on the confidence of our teacher model keeping only high-confident samples while maintaining the ratio of labels of the training set of the downstream task. For relatively small tasks, we use a threshold such that our augmented training set is approximately a hundred times bigger, and for datasets of medium size, only ten times bigger.

\subsection{Semi-supervised learning for natural language understanding}

We combine our data augmentation technique with self-training and knowledge distillation, two semi-supervised learning techniques that benefit from having relevant unannotated sentences. 

\insertdatasets

\paragraph{Self-training.}
Following the steps in \autoref{fig:sentaugment}, we first train a teacher model by fine-tuning a pretrained RoBERTa-Large model on the target downstream task. We then use it to annotate the retrieved in-domain sentences. For each class, we select the sentences with the highest scores and prune the rest. We make sure the label ratio is maintained between the original downstream task training set and the augmented set by considering the probability of the classifier. As our student model, we then finetune a new RoBERTa-Large using KL-divergence on the synthetic data by considering the post-softmax class probabilities as labels. 

\paragraph{Knowledge-distillation.}
We follow the same approach for knowledge-distillation, except
we consider a student model that has an order of magnitude less parameters than the RoBERTa-Large teacher model. As for self-training, we pretrain the student and use continuous probabilities as synthetic labels. We exploit data augmentation by using in-domain unannotated sentences.

\paragraph{Few-shot learning.}
Semi-supervised learning techniques are adapted to settings where little supervised data is available. We simulate a few-shot learning environment by only considering a few samples per class, for several downstream tasks. We apply data augmentation and self-training in that context by augmenting the training set by two to three orders of magnitude more data and use a teacher model trained on only a few training samples to synthetically annotate data.

\section{Experimental setup}
Next, we give details on how we build the bank of sentences, what downstream tasks we use for evaluation and we describe our training procedure for semi-supervised learning.

\subsection{Large-scale bank of sentences}
As a large-scale external bank of unannotated sentences, we extract and filter text from CommonCrawl~\footnote{\url{www.github.com/facebookresearch/cc_net}}~\citep{wenzek2019ccnet}. In particular, we apply a simple sentence segmenter to turn documents into sentences and perform deduplication. We refer to samples in this dataset as sentences although is also contains shorts spans of text that can be seen as short documents. We use three corpora, CC-100M with one hundred million sentences (2B words), CC-1B with one billion sentences (20B words) and CC-5B with five billion sentences (100B words), the first two being random subsets of the biggest one. 
When retrieving sentences, we remove those that overlap with sentences from the test set of the downstream task. CommonCrawl data contains a wide variety of domains and text styles which makes it a good general-purpose corpus. We will release our code to obtain a similar corpus.

\subsection{Evaluation datasets}
We evaluate our approach on the Stanford Sentiment Treebank~\citep{socher2013recursive} binary and fine-grained sentiment analysis datasets (SST-2 and SST-5), on product classification (CR) from \cite{Hu:2004kdd}, hate-speech comment classification\footnote{\url{www.kaggle.com/c/detecting-insults-in-social-commentary/overview}} (IMP), question classification (TREC) from \cite{Voorhees:2000sigir} and named entity recognition (CoNLL 2002) from \cite{sang2003introduction}. We provide details of each task including task, domain, size and number of classes in Table~\ref{tab:datasets}.

\insertselftrain
\insertfewshot

\subsection{Training details}

\paragraph{Our sentence embeddings.}
We train our own SentAugment Sentence Encoder (SASE) by leveraging paraphrases from NLI entailment pairs~\citep{williams2017broad}, MRPC~\citep{dolan2005automatically}, Quora Question Pairs (QQP), round-trip translation~\citep{wieting2017paranmt} and web paraphrases~\citep{creutz2018open}, together with OpenSubtitles~\citep{lison2019open} and Europarl~\citep{koehn2005europarl} parallel data from English to French, Italian and Indonesian - language pairs that were shown to provide good paraphrastic sentence embeddings~\citep{wieting2019simple}. We pretrain the model with a multilingual masked language modeling objective~\citep{devlin2018bert,conneau2019cross} in these 4 languages, with a sentence piece segmentation trained on a corpus with 3/4 of English data to give more importance to English, and the rest in other languages. We use a triplet loss to learn cosine sentence embedding similarity where the negative is selected to be the hardest in the batch.
We evaluate our model on STS benchmarks~\citep{Agirre:2012semeval} and report results in Section~\ref{sec:sase} where we show our model outperforms previous approaches. We found that due to pretraining and being trained on longer sentences, our model is also more adapted to raw and long sentences from CommonCrawl. We also consider word2vec embeddings~\citep{mikolov2013word2vec} and the uSIF approach~\citep{ethayarajh2018repnlp,arora2019simple} as baselines in our experimental results. 

\paragraph{Fine-tuning the student model.}
We use fairseq~\citep{ott2019fairseq} and the open-source RoBERTa-Large model~\citep{roberta2019} as our pretrained Transformer baseline and perform finetuning on each downstream task. We use Adam, with learning-rate schedule 1e-5. We use batch-sizes of 16 and dropout rate 0.1. We fine-tune on synthetically annotated data using KL divergence. We found that fine-tuning again on the training set of the downstream task with ground-truth labels was not necessary, neither was adding ground-truth sentences from the training set to the self-training data.

\insertkd

\paragraph{Few-shot learning experiments.} We sample 5 training sets that each consist of 20 examples from each label from the original training set of the task. We sample 200 examples from the original validation set of the task, taking the label distribution into account. We use the original test set of the task as our test set. For all experiments, we run 10 seeds for each train set and consider the mean test accuracy of top 3 models (based on their validation accuracy) as the performance on that train set. Based on this, we calculate the mean and standard deviation across 5 training sets, to report our final results. We synthetically annotate both retrieved and ground-truth data, and train each model for 50 epochs. Different from our experiments in the full-shot setting, we (1) use discrete labels, (2) include ground truth data in the training set, and (3) augment the reduced training set by one order of magnitude data samples sampled from the top 1000*(total supervised examples).
These choices were made for few-shot learning experiments as the teacher model is not as strong, leading to noisier annotations compared to the full dataset setup.

\section{Analysis and Results}
\label{sec:analysis}
In this section, we first report results on self-training, knowledge-distillation and few-shot learning with our best approach. We then provide an analysis of the key factors that makes self-training with SentAugment work in the context of natural language understanding.
%, as well as important components of SentAugment.

\subsection{Self-training experiments}
In Table ~\ref{tab:selftrain}, we report results using self-training on six different downstream tasks. To understand the contribution of domain-adaptation and the actual contribution of self-training (ST), we compare ST to in-domain continued pretraining (ICP) where we continue masked language model pretraining of a RoBERTa-Large model on the retrieved in-domain augmented data. The goal of this comparison is to understand whether self-training only does domain adaptation to the target domain of the downstream task, which ICP also does. Indeed, RoBERTa-Large has been trained on a very large generic dataset of web data but not particularly specific to each downstream task.

First, we observe that self-training alone improves performance over a strong RoBERTa-Large baseline, leading to an 1.2\% improvement on average. Improvements are largest on SST-5 and IMP, with 2.6\% and 3.1\% improvements respectively. On the other hand, when continuing pretraining on the self-training data with ICP, we observe a decrease in performance from 87.4\% to 86.2\%. It is interesting to note that this is not only the use of the in-domain data that is useful but the combination with the self-training algorithm. While ICP performs domain adaptation at pretraining time of the RoBERTa-Large model, it does not outperform the baseline. Self-training is thus a nontrivial way of improving generalization and doing domain-adaptation at fine-tuning time. \cite{xie2019unsupervised} however show gains using ICP. We attribute that difference in our conclusion to (i) RoBERTa being trained on much more data than their BERT model trained on Wikipedia, (ii) our ICP using only approximately in-domain data rather than ground-truth.

\subsection{Few-shot learning experiments}

We investigate the effectiveness of our approach in the context of few-shot learning. In Table ~\ref{tab:fewshot}, we fine-tune a RoBERTa-Large model on between 40-200 samples of training data in each task and use it as a teacher model. Self-training leads to 3.5\% average gains on all tasks, going from 72.0\% to 75.5\% while also reducing the variance. Gains are particularly strong on sequence labeling, where the student model obtains 58.4 F1 over 49.0 F1 for the teacher model.

\subsection{Knowledge distillation experiments}
Knowledge distillation (KD) also strongly benefits from large-scale augmentation. Table ~\ref{tab:kd} shows baseline results from the RoBERTa-Large and RoBERTa-Small directly fine-tuned on the training set of each downstream task. Comparing distilled models that use different kinds of unannotated data, we observe that using the ground-truth (GT) leads to significantly better performance compared to random (RD) sentences, going from 77.1\% to 82.5\%. This shows that assuming the existence of data in the exact same domain is a strong assumption. Using the same amount of data, our data augmentation (SA) method bridges the gap with 81.9\% average accuracy.

When leveraging more unannotated sentences, we push the random baseline to 82.1\% which corresponds to a 5\% improvement, getting closer to the GT baseline. Finally, using SentAugment leads to strong improvements, up to 85.4\% average accuracy, only 0.9\% average accuracy below the teacher model with almost ten times less parameters, showing the importance of data augmentation for KD.

\subsection{Ablation study of data augmentation}
Our approach leverages several key components that make data augmentation work and that enable self-training for natural language understanding. We examine these components in this section.

\paragraph{Task-specific retrieval.}
We compare different methods for building task-specific embeddings used as queries for retrieving in-domain sentences from the large bank of sentences. In Table~\ref{tab:augmentation}, we observe that using one query for each label (label-average) leads to better performance than having a single query embedding for the entire task (all-average), leading to a 83.1\% accuracy on average. For tasks with unbalanced classes, this avoids an over-representation of the majority class, and also provides more diversity in the retrieved sentences. Interestingly, having one query embedding per sentence in the training set does not improve performance, except for named entity recognition where the per-sentence approach leads to the best performance.
\insertaugmentation

\paragraph{Sentence embedding space.}
Our data augmentation method is based on structuring a large external bank of text with a sentence embedding space. The sentence embedding method plays an essential role as shown in Table~\ref{tab:embeddings}. We compare three embedding methods, the average of fastText~\citep{mikolov2018advances} word embeddings (average-word2vec), the uSIF-ParaNMT embeddings~\citep{ethayarajh2018repnlp} and our own sentence encoder. %We refer to Appendix~\ref{sec:appendix_A} for more details on the sentence encoders and their results on semantic textual similarity tasks. 
We observe that uSIF-ParaNMT and para-embeddings - two sentence embedding methods that obtain state-of-the-art results on semantic textual similarity benchmarks - lead to stronger performance than the average-word2vec approach. Para-embeddings leads to the best performance and improves performance over uSIF by 0.4\% on average.
\insertembedding

\paragraph{Scaling bank size.}
To demonstrate the importance of large-scale retrieval, we evaluate our method using an increasing amount of data for our bank, from fifty million sentences to five billion sentences (one hundred billion words). We observe a significant increase in performance from 50m to 1B in Table~\ref{tab:memory}, but the improvement seems to saturate when going from 1B to 5B. 
However, the 5B external bank may however provide additional gains for tasks that are in rare domains and that can leverage the additional 4B sentences, which correspond to 342M additional CommonCrawl documents. Another effect of increasing the corpus size may be reducing diversity in the retrieved sentences. We leave experimenting with diversity-inducing enhancements to the retrieval for future work.

\insertmemory

\paragraph{Continuous labels.}
In Table~\ref{tab:logits}, we show that using class probabilities as synthetic labels leads to significantly better performance, outperforming discrete synthetic labels by 0.9\% on average. We found very little gain when using self-training with discrete labels, contrary to previously published results in computer vision~\citep{yalniz2019billion,xie2020self}. A difference with previous work in computer vision is the number of classes of the supervised data. In that context, discrete labels provide even less information to the student model than continuous class probabilities.

\insertlabel
\insertanalysis

\paragraph{Computational cost of self-training.}
SentAugment data prefiltering reduces the amount of data to be annotated by the teacher model and also filters based on the target domain. Filtering based solely on classifier confidence is significantly more expensive computationally, as annotating 10000 sentences with RoBERTa-Large takes approximately 3 seconds on a Volta-32GB GPU. This means that annotating 1B sentences takes 83 hours on a single GPU and much longer for models of larger size such as T5~\citep{raffel2019exploring} or GPT-3~\citep{brown2020language}. On the other hand, using SentAugment based on a few task-specific query embedding (label-average) takes one minute for scoring 1B sentences. By only selecting the first few million top sentences, or less, to synthetically annotate, this greatly reduces computational cost and allows to scale to a larger bank of sentences, which in turn allows for more domains to be considered. 
Note that similarity search can be further sped up significantly by using fast nearest neighbor search such as product quantization with inverted files~\citep{johnson2019billion}.

\section{Analysis of similarity search}
\label{sec:sase}
In this section, we present the results of our SentAugment sentence embedding (SASE) method on semantic textual similarity (STS) benchmarks and present examples of retrieved sentence based on large-scale similarity search.

\subsection{Sentence embeddings (SASE)}
In Table~\ref{tab:sts}, we compare our sentence embedding method to previous approaches including BERT (Mean)~\cite{devlin2018bert}, InferSent~\cite{conneau-EtAl:2017:EMNLP2017}, GenSen~\cite{subramanian2018learning}, USE~\cite{cer2018universal}, Sentence-BERT~\cite{reimers2019sentence}, uSIF~\cite{ethayarajh2018unsupervised}, Charagram~\cite{wieting2017paranmt} and BGT~\cite{wieting2019bilingual}. On average, our embeddings outperform previous approaches by 0.2\% on STS 2012 to 2016~\cite{Agirre:2012semeval,Agirre13sem2013,agirrea2014semeval,agirre2015semeval,agirrea2016semeval}, and by 0.9\% on STS-Benchmark~\cite{cer2017semeval}.
\insertsts

\subsection{Examples of large-scale similarity search}
SentAugment uses large-scale similarity search combined with an embedding space with billions of sentences to find in-domain sentences. In Table ~\ref{tab:analysis}, we show examples of nearest neighbors extracted from CommonCrawl based on sentence-level or label-level queries and for different domains such as biomedical, financial or hate-speech data. We see that retrieving nearest neighbors can lead to good paraphrases which either preserve the meaning or augment it with additional information. We also observe reformulation of the same input sentence. As for label-level queries, we observe that retrieved sentences match very well the domain of the downstream task.
In the Appendix, we provide more examples of how well the model performs when trying to find similar sentences in our corpus using our sentence embedding. We will release our indexes for people to explore further large-scale similarity search of web data. We hope this will lead to an improved understanding of large-scale sentence embedding spaces but also help the community analyze the content and biases of large-scale web corpora used to train language models.

\section{Conclusion}
Recent work in natural language understanding has focused on unsupervised pretraining. 
In this paper, we show that self-training is another effective method to leverage unlabeled data. 
We introduce SentAugment, a new data augmentation method for NLP that retrieves relevant sentences from a large web data corpus. 
Self-training is complementary to unsupervised pre-training for a range of natural language tasks and their combination leads to further improvements on top of a strong RoBERTa baseline.
We also explore knowledge distillation and extend previous work on few-shot learning by showing that open domain data with SentAugment is sufficient for good accuracy.

% \section*{Acknowledgements}

\clearpage
\bibliography{acl2020}
\bibliographystyle{acl_natbib}

%%%%%%%%%%%%%%%%%%%%%%%%%%%%%%%%%%%%%%%%%%%%%%%%%%%%%%%%%%%%%%%%%%%%%%%%%%%%%%%
%%%%%%%%%%%%%%%%%%%%%%%%%%%%%%%%%%%%%%%%%%%%%%%%%%%%%%%%%%%%%%%%%%%%%%%%%%%%%%%
% DELETE THIS PART. DO NOT PLACE CONTENT AFTER THE REFERENCES!
%%%%%%%%%%%%%%%%%%%%%%%%%%%%%%%%%%%%%%%%%%%%%%%%%%%%%%%%%%%%%%%%%%%%%%%%%%%%%%%
%%%%%%%%%%%%%%%%%%%%%%%%%%%%%%%%%%%%%%%%%%%%%%%%%%%%%%%%%%%%%%%%%%%%%%%%%%%%%%%
% \newpage
% \clearpage
% \appendix
% \section{Appendix - similarity search examples}
% In Table ~\ref{tab:examples}, we show examples of nearest neighbors retrieved using similarity search. We use as queries the label-average embeddings of each downstream task, for different classes.
% \insertexamples

% In Table~\ref{tab:sts}, we report results on STS benchmarks from 2012 to 2016 as well as the STS benchmark results from previous work and our own sentence encoder. We reaport the Pearson correlation.
% \insertsts
\end{document}

%% file: content/tables.tex
\newcommand{\insertapproach}{
\begin{figure*}[t]
	\begin{center}
        \includegraphics[width=\linewidth]{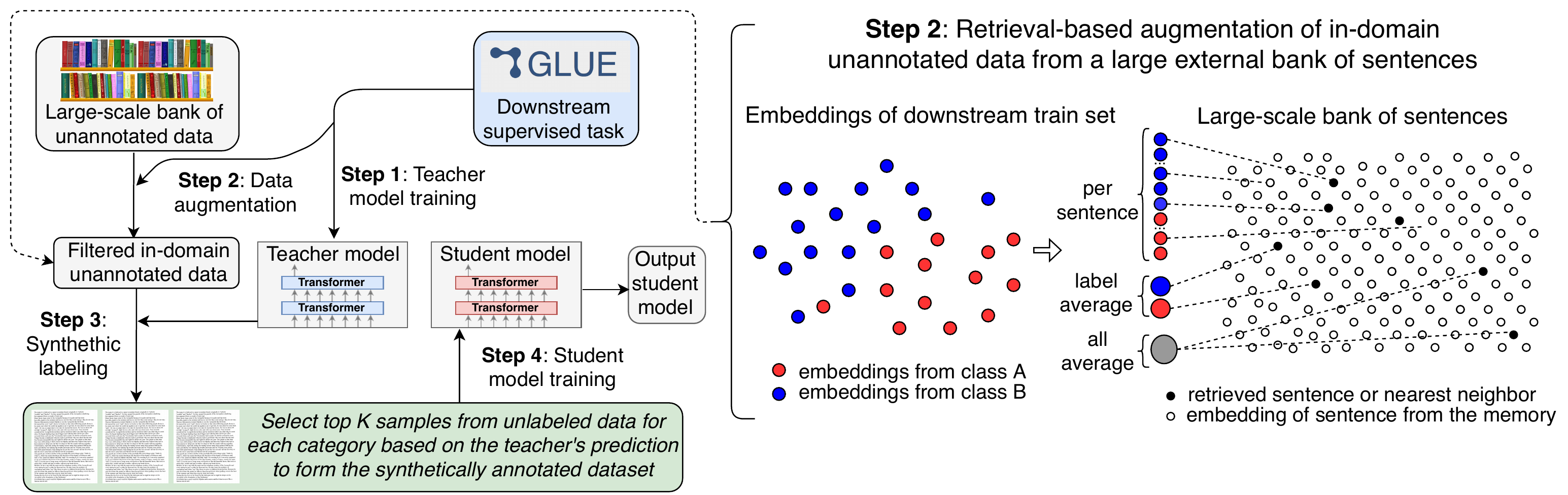}
        \caption{\textbf{The SentAugment approach.} The self-training procedure follows multiple steps; \textbf{Step 1}: A RoBERTa-Large model - the teacher - is finetuned on a downstream task using a cross-entropy loss, \textbf{Step 2}: Task-specific unannotated data is extracted from a large bank of sentences; This step uses task-specific query embeddings (produced by a paraphrastic sentence encoder) to select nearest neighbors from the bank. \textbf{Step 3}: This data is synthetically annotated using the teacher model; top K samples from each class are selected to form the final synthetic dataset; \textbf{Step 4}: A RoBERTa-Large model - the student - is finetuned on this dataset using KL-divergence. Our approach differs from previous work at Step 2, which we show is crucial for open-domain self-training.
            \label{fig:sentaugment}}
        \vspace{-0.2cm}
	\end{center}
\end{figure*}
}

\newcommand{\insertselftrain}{
    \begin{table*}[h!]
        \begin{center}
            \resizebox{0.8\linewidth}{!}{
            \begin{tabular}[b]{lccccccccc}
            \toprule
                {\bf Model} & {\bf SST-2} & {\bf SST-5} & {\bf CR} & {\bf IMP} & {\bf TREC} & {\bf NER} & {\bf Avg} \\
                \midrule
                \midrule
                RoBERTa$_{\text{Large}}$ & 96.5 & 57.8 & 94.8 & 84.6 & \bf 97.8  & 92.7 & 87.4 \\
                RoBERTa$_{\text{Large}}$ + ICP & 93.9 & 55.1 & 93.7 & 84.4 & 97.8  & 92.1 & 86.2 \\
                RoBERTa$_{\text{Large}}$ + ST & \bf 96.7 & \bf 60.4 & \bf 95.7 & \bf 87.7 & \bf 97.8  & \bf 93.3 & \bf 88.6 \\
                \bottomrule
            \end{tabular}
            }
            \caption{Results of self-training on natural language understanding benchmarks. We report a strong RoBERTa-Large baseline, as well as in-domain continued pretraining of this model (ICP) and our self-training approach (ST).
            \label{tab:selftrain}}
        \end{center}
       \vspace{-0.4cm}
    \end{table*}
}

\newcommand{\insertfewshot}{
    \begin{table*}[h]
        \begin{center}
            \resizebox{\linewidth}{!}{
            \begin{tabular}[t]{lccccccc}
            \toprule
                {\bf Model} &  {\bf SST-2} & {\bf SST-5} & {\bf CR} & {\bf IMP} & {\bf TREC} & {\bf NER} & {\bf Avg} \\
                \midrule
                \midrule
                \multicolumn{1}{l}{Num samples} & 40 & 100 & 40 & 40 & 120 & 200 & - \\
                \midrule
                \midrule
                RoBERTa$_{\text{Large}}$ & 83.6$\pm$2.7 & 42.3$\pm$1.6 & 88.9$\pm$1.7 & 77.3$\pm$2.8 & 90.9$\pm$2.5 & 49.0$\pm$1.7 & 72.0$\pm$2.2\\
                RoBERTa$_{\text{Large}}$ + ST & \textbf{86.7$\pm$2.3} & \textbf{44.4$\pm$1.0} & \textbf{89.7$\pm$2.0} & \textbf{81.9$\pm$1.4} & \textbf{92.1$\pm$2.4} & \textbf{58.4$\pm$1.4} & \textbf{75.5$\pm$1.8} \\
                \bottomrule
            \end{tabular}
            }
            \caption{Results of self-training for few-shot learning, using only 20 samples per class.
            \label{tab:fewshot}}
        \end{center}
       \vspace{-0.4cm}
    \end{table*}
}

\newcommand{\insertkd}{
    \begin{table*}[h!]
        \begin{center}
            \resizebox{0.85\linewidth}{!}{
            \begin{tabular}[t]{lccccccc}
            \toprule
                {\bf Model} & {\bf KD-data} & {\bf SST-2} & {\bf SST-5} & {\bf CR} & {\bf IMP} & {\bf TREC} & {\bf Avg} \\
                \midrule
                \midrule
                \multicolumn{8}{l}{\it Models trained directly on the training set of each downstream task} \\
                \midrule
                \midrule
                RoBERTa$_{\text{Large}}$ & - & \bf 96.5 & \bf  57.8 & \bf 94.8 & \bf 84.6 & \bf 97.8 & \bf 86.3 \\
                RoBERTa$_{\text{Small}}$ & - & 92.0 & 49.0 & 88.7 & 83.8 & 96.4 & 82.0 \\
                \midrule
                \midrule
                \multicolumn{8}{l}{\it Models distilled using the same number of sentences as in the train set (cf. Table~\ref{tab:datasets})} \\
                \midrule
                \midrule
                RoBERTa$_{\text{Small(Large)}}$ & GT & 92.4& 49.7 & 89.6 & 84.4 & 96.6 & 82.5 \\
                RoBERTa$_{\text{Small(Large)}}$ & RD & 90.7 & 47.5 & 87.4 & 69.1 & 90.8 & 77.1 \\
                RoBERTa$_{\text{Small(Large)}}$ & SA & 91.8 & 50.7 & 88.2 & 84.6 & 94.4 & 81.9 \\
                \midrule
                \midrule
                \multicolumn{8}{l}{\it Models distilled using more unsupervised sentences (100k sentences)} \\
                \midrule
                \midrule
                RoBERTa$_{\text{Small(Large)}}$ & RD & 92.5 & 51.2 & 92.4 & 78.1 & 96.2 & 82.1 \\
                RoBERTa$_{\text{Small(Large)}}$ & SA & \bf 94.2 & \bf 57.6 & \bf 92.6 & \bf 85.5 & \bf 97.0 & \bf 85.4 \\
                \bottomrule
            \end{tabular}
            }
            \caption{Results of knowledge-distillation using ground-truth (GT), random (RD), or data-selected data (SA) as unnanotated sentences. We distill a RoBERTa-Large model of 24 layers into a RoBERTa-Small model with $100\times$ less parameters. \label{tab:kd}}
        \end{center}
       \vspace{-0.4cm}
    \end{table*}
}

\newcommand{\insertaugmentation}{
    \begin{table}[h!]
        \begin{center}
            \resizebox{1\linewidth}{!}{
            \begin{tabular}[h!]{llccccc}
            \toprule
                {\bf Model} & {\bf Selection} & {\bf $\mathcal{C}$} & {\bf SST-5} & {\bf CR} & {\bf NER} & {\bf Avg} \\
                \midrule
                RoBERTa$_{\text{Large}}$ + ST & all-avg & $\mathcal{O}(Md^2)$ & 60.0 & 94.7 & 92.8 & 82.5 \\
                RoBERTa$_{\text{Large}}$ + ST & label-avg & $\mathcal{O}(KMd^2)$ & \bf 60.4 & \bf 95.7 & 93.1 & \bf 83.1 \\
                RoBERTa$_{\text{Large}}$ + ST & per-sent & $\mathcal{O}(NMd^2)$ & 60.1 & 95.4 & \bf 93.3 & 82.9 \\
                \bottomrule
            \end{tabular}
            }
            \caption{Impact of data augmentation technique. $\mathcal{C}$ is the complexity, $M$ the size of the bank of sentences, $K$ the number of labels (or clusters), $N$ the size of the downstream training set and $d$ the embedding size.\label{tab:augmentation}}
        \end{center}
       \vspace{-0.4cm}
    \end{table}
}

\newcommand{\insertembedding}{
    \begin{table}[h!]
        \begin{center}
            \resizebox{1\linewidth}{!}{
            \begin{tabular}[b]{lcccccc}
            \toprule
                {\bf Model} & {\bf Embedding} & {\bf dim} & {\bf SST-5} & {\bf CR} & {\bf NER} & {\bf Avg} \\
                \midrule
                RoBERTa$_{\text{Large}}$ + ST & avg-w2v & 300 & 59.4 & 95.2 & 92.9 & 82.5 \\
                RoBERTa$_{\text{Large}}$ + ST & uSIF & 300 & 59.9 & 95.0 & 93.1 & 82.7 \\
                RoBERTa$_{\text{Large}}$ + ST & SASE & 256 & \bf 60.4 & \bf 95.7 & \bf 93.1 & \bf 83.1 \\
                \bottomrule
            \end{tabular}
            }
            \caption{Impact of sentence embedding method: average-word2vec, uSIF with ParaNMT and SASE. \label{tab:embeddings}}
        \end{center}
       \vspace{-0.4cm}
    \end{table}
}

\newcommand{\insertmemory}{
    \begin{table}[h!]
        \begin{center}
            \resizebox{1\linewidth}{!}{
            \begin{tabular}[b]{lcccccc}
            \toprule
                {\bf Model} & {\bf \#lines} & {\bf \#words} & {\bf SST-5} & {\bf CR} & {\bf NER} & {\bf Avg}  \\
                \midrule
                RoBERTa$_{\text{Large}}$ + ST & 50m & 1B & 59.5 & 95.4 & 92.8 & 82.6  \\
                RoBERTa$_{\text{Large}}$ + ST & 250m & 5B & 59.5 & \bf 95.7 & 92.9 & 82.7  \\
                RoBERTa$_{\text{Large}}$ + ST & 1B & 20B & \bf 60.4 & \bf 95.7 & \bf 93.1 & \bf 83.1  \\
                RoBERTa$_{\text{Large}}$ + ST & 5B & 100B & 60.0 & 95.3 & \bf 93.1 & 82.8  \\
                \bottomrule
            \end{tabular}
            }
            \caption{Impact of sentence bank size (number of lines and words) on self-training results. \label{tab:memory}}
        \end{center}
       \vspace{-0.4cm}
    \end{table}
}

\newcommand{\insertdatasets}{
    \begin{table*}[t]
        \begin{center}
            \resizebox{0.9\linewidth}{!}{
            \begin{tabular}[b]{lllrc}
            \toprule
                {\bf Dataset} & {\bf task} & {\bf domain} & {\bf \#train} & {\bf \#classes}\\
                \midrule
                SST-2 & sentiment analysis & movie reviews & 67349 & 2 \\
                SST-5 & sentiment analysis & movie reviews & 8544 & 5\\
                CR & product classification & product reviews & 2500 & 2  \\
                IMP & hate-speech classification & forum conversations & 3947 & 2 \\
                TREC & question-type classification & short questions & 5001 & 6 \\
                CoNLL & named entity recognition & news stories & 11663 & 5  \\

                \bottomrule
            \end{tabular}
            }
            \caption{Downstream tasks used for evaluation. \label{tab:datasets}}
        \end{center}
       \vspace{-0.4cm}
    \end{table*}
}

\newcommand{\insertlabel}{
    \begin{table}[h!]
        \begin{center}
            \resizebox{1\linewidth}{!}{
            \begin{tabular}[b]{lccccc}
            \toprule
                {\bf Model} & {\bf label type} & {\bf SST-5} & {\bf CR} & {\bf NER} & {\bf Avg} \\
                \midrule
                RoBERTa$_{\text{Large}}$ + ST & discrete & 59.1 & 94.7 & 92.8  & 82.2  \\
                RoBERTa$_{\text{Large}}$ + ST & logits &  \bf 60.4 & \bf 95.7 & \bf 93.1  & \bf 83.1 \\
                \bottomrule
            \end{tabular}
            }
            \caption{Impact of label type on self-training results. \label{tab:logits}}
        \end{center}
       \vspace{-0.4cm}
    \end{table}
}

\newcommand{\insertanalysis}{
    \begin{table*}[h!]
    \resizebox{1\linewidth}{!}{
        \begin{tabular}{l}
        \toprule
        \textbf{BioNLP query}: A single gene on chromosome 7 makes a protein called the cystic fibrosis transmembrane conductance regulator (CFTR).	 \\
        \textbf{Nearest neighbor}: Cystic Fibrosis A mutation in the gene cystic fibrosis transmembrane conductance regulator (CFTR) in chromosome 7. \\
        \midrule
        \textbf{Financial Query}: Google has entered into an agreement to buy Nest Labs for \$3.2 billion. \\
        \textbf{Nearest neighbor}: In January Google (NASDAQ:GOOG) reached an agreement to buy Nest Labs for \$3.2 billion in cash.\\
        \midrule
        \textbf{Hate-speech Query}: \textit{Average sentence embeddings of the "hateful" class of IMP} \\
        \textbf{Nearest neighbor}: fuzzy you are such a d* f* piece of s* just s* your g* d* mouth. -- All you n* and s* are fucking ret*\\
        \midrule
        \textbf{Movie review Query}: \textit{Average sentence embeddings of the "bad movie" class of SST-5} \\
        \textbf{Nearest neighbor}: This movie was terribly boring, but so forgettable as well that it didn't stand out for how awful it was.. \\
        \midrule
        \textbf{Product review Query}: \textit{Average sentence embeddings of the "positive" class of CR} \\
        \textbf{Nearest neighbor}: The phone is very good looking with superb camera setup and very lightweight. \\
        \midrule
        \textbf{Question type Query}: \textit{Average sentence embeddings of the "location" class of TREC} \\
        \textbf{Nearest neighbor}: Lansing is the capital city of which state? \\
        \bottomrule
        \end{tabular}
        \smallskip
    }
    \caption{\small Examples of nearest neighbors using a per-sentence or label-average query from different domains.
    \label{tab:analysis}}
    \end{table*}
}

\newcommand{\insertsts}{
    \begin{table}[h!]
        \begin{center}
            \resizebox{1\linewidth}{!}{
            \begin{tabular}[b]{l|ccccc|c|c}
            \toprule
                \multirow{2}{*}{{Model}} & \multicolumn{6}{c|}{Semantic Textual Similarity (STS)} &  \\
                 & 2012 & 2013 & 2014 & 2015 & 2016 & Avg & STS-B \\
                \midrule
                BERT (Mean) & 48.8 & 46.5 & 54.0 & 59.2 & 63.4 & 54.4 & - \\
                InferSent & 61.1 & 51.4 & 68.1 & 70.9 & 70.7 & 64.4 & 70.6  \\
                GenSen & 60.7 & 50.8 & 64.1 & 73.3 & 66.0 & 63.0 & -  \\
                USE & 61.4 & 59.0 & 70.6 & 74.3 & 73.9 & 67.8 & -  \\
                Sentence-BERT & 66.9 & 63.2 & 74.2 & 77.3 & 72.8 & 70.9 & -  \\
                uSIF- & 68.3 & {\bf 66.1} & {\bf 78.4} & 79.0 & - & - & 79.5  \\
                Word, trigram & 67.8 & 62.7 & 77.4 & 80.3 & 78.1 & 73.3 & 79.9  \\
                BGT & 68.9 & 62.2 & 75.9 & 79.4 & {\bf 79.3} & 73.1 & -  \\
                \midrule
                SASE (ours) & {\bf 69.7} & 62.9 & 77.3 & {\bf 79.8} & 78.1 & {\bf 73.5} & {\bf 80.8}  \\
                \bottomrule
            \end{tabular}
            }
            \caption{Results of our sentence encoder (SASE) on STS benchmarks from 2012 to 2016 and on the test sets of the STS-Benchmark dataset, compared to previously published results. We report Pearson’s r $\times$ 100. \label{tab:sts}}
        \end{center}
       \vspace{-0.4cm}
    \end{table}
}